# An Algorithm for Fuzzification of WordNets, Supported by a Mathematical Proof

Sayyed-Ali Hossayni [a,b,c], Mohammad-R Akbarzadeh-T [a] Diego Reforgiato Recupero [d,f], Aldo Gangemi [e,d,g],
Esteve Del Acebo [c] and Josep Lluís de la Rosa i Esteva [c]

*Abstract*— WordNet-like Lexical Databases (WLDs) group English words into sets of synonyms called "synsets." Although the standard WLDs are being used in many successful Text-Mining applications, they have the limitation that word-senses are considered to represent the meaning associated to their corresponding synsets, to the same degree, which is not generally true. In order to overcome this limitation, several fuzzy versions of synsets have been proposed. A common trait of these studies is that, to the best of our knowledge, they do not aim to produce fuzzified versions of the existing WLD's, but build new WLDs from scratch, which has limited the attention received from the Text-Mining community, many of whose resources and applications are based on the existing WLDs. In this study, we present an algorithm for constructing fuzzy versions of WLDs of any language, given a corpus of documents and a word-sense disambiguation (WSD) system for that language. Then, using the Open-American-National-Corpus and UKB WSD as algorithm inputs, we construct and publish online the fuzzified version of English WordNet (FWN). We also propose a theoretical / mathematical proof of the validity of its results.

## I. INTRODUCTION

In 1990, Miller et al. [1] proposed WordNet (WN) [2][3], a lexical database for the English language that groups English words into synonym sets, called synsets[1]. From there on, based on the WN structure, other lexical databases were also proposed for different languages [4][5][6] that collect synsets of their corresponding languages, as it is done in WN. We call these lexical databases under the umbrella-term WordNet-like Lexical Database (WLD). WLDs have a wide variety of applications in Natural Language Processing [7][8], Knowledge Engineering [9][10], and Ontology Engineering [11][12].

However, in WLDs, all the members of a synset are supposed to belong to a synset with the same degree and convey the meaning of that synset at the same level. In other words, WLDs assume synsets to be crisp (non-fuzzy) sets. But this simple assumption does not always properly model the complex nature of meaning in natural languages. For example, let's consider the following synset of WN: *Synset('flower.n.02'): {flower, bloom, blossom};* it contains the words that potentially (as one of their senses) stand for "reproductive organ of angiosperm plants especially one having showy or colorful parts" (the illustrative-definition of each synset is proposed in WN).

Before proceeding with the mentioned issue, it is worthy to introduce the concept of a "lemma" and the concept of a "word-sense," in WLDs: (1) Each word disregarding its various potential senses is called a "lemma". For example, "bloom" disregarding the sense for which it can stand is considered a lemma. It is also the case for all the words of a dictionary. (2) A specific sense of a lemma that is logically a member of one specific synset, is called a word-sense. For example, the above-mentioned sense of the lemma "bloom" is called a word-sense[2].

Usually, the lemmas (e.g. flower, bloom …) related to the word-senses of a synset (e.g. Synset('flower.n.02')), are not equally compatible with the meaning (definition) of the synset, and each of them can have a different degree of compatibility. Therefore, the concept of fuzzy synsets was proposed. Since 2005, some researches are being conducted, studying on fuzzy synsets and the resulting WLDs.

In 2005, Veldall [13], without using the term "fuzzy synset" (even without using the term "synset"), proposed an algorithm for creating fuzzy semantic classes[3] (i.e. synsets) and stated that "different words can represent more or less typical instances of a given concept. Some words may represent clear-cut instances of a given category, while others represent peripheral or border-line cases." In order to illustrate such categories, they considered them as fuzzy sets, utilized a fuzzy clustering algorithm for assigning membership values of the corresponding members, and proposed a Norwegian fuzzy WLD. In 2010, Borin and Forsberg [14] who (to the best of our knowledge) coined the term "fuzzy synsets," viewed them from a pure linguistics point of view, and based them on "synonymy avoidance" [15] which implies that two word-senses of a human language are very unlikely to exactly stand

This research has partly been supported by AGAUR res. grant 2013 DI 012, AHD 13960201-01 and *-02, QWAVES- 2014-2576-7, ANSwER- RTC-2015-4303-7; CSI-ref. 2017 SGR 1648.

The corresponding author (SA Hossayni) has conducted this paper during his PhD course in Universitat de Girona (Spain). However, most of this research has been performed while his official sabbatical in SCIIP, STLAB, and LIPN:

[a] SCIIP, Dep. of Comp. Eng., Ferdowsi University of Mashhad, Mashhad, Iran.
[b] CogLing, DM Lab, Dep. of Comp. Eng., Iran Univ. Sci. Tech., Tehran, Iran.
[c] ARLab, Centre EASY, ViCOROB, Montilivi Campus, Univ. Girona, Spain.
[d] STLab, ISTC, Italian National Research Council (CNR), Italy
[e] LIPN, Université Paris 13 - Sorbonne Paris Cité - CNRS, France
[f] Dep. Math. and Comp. Sci., University of Cagliari, Italy
[g] Dep. Classical Philology and Italian Studies, University of Bologna, Italy

hossayni@iran.ir; akbazar@um.ac.ir; diego.reforgiato@unica.it; aldo.gangemi@unibo.it ; esteve.acebo@udg.edu; peplluis@silver.udg.edu

[1] It additionally providing short definitions and usage examples and records a number of relations among these synsets and their members.

[2] Each lemma can have several word-senses. In other words, each lemma can be a member of more than one synset.

[3] He applied his algorithm on Norwegian language.

for a same meaning / definition. Consequently, a dictionary that fundamentally assumes synonymy (linguistically speaking) cannot fairly project human lexical knowledge. In the mentioned study, Borin et al. [14][16] utilize Synlex (People's synonym lexicon [17] that contains synonymy[4] degree of word-pairs, provided by crowdsourcing) as well as SALDO[5] [18][19] to present an algorithm to create fuzzy synsets for the Swedish language. In 2011, Gonçalo and Gomes [20] were the second research group which looked at fuzzy synsets from a linguistics point of view expressing that "from a linguistic point of view, word senses are not discrete and cannot be separated with clear boundaries [21] [22][6]… Sense division in dictionaries and lexical resources is most of the times artificial…" They proposed an algorithm for generating fuzzy synsets and applied it to the Portuguese language, producing a Portuguese fuzzy WLD.

However, to the best of our knowledge, none of the mentioned studies, directed towards a fuzzy understanding of synsets, have proposed any approach to produce a fuzzy version of the crisp synsets in the existing WLDs (e.g. WordNet, EuroWordNet, Arabic WordNet, IndoWordNet …). In other words, in the mentioned few studies, the synsets either are not predefined and can be determined only after running the proposed algorithm (i.e. fuzzy synsets are the output of clustering [13][20]), or there exists a lexical database (SALDO in [14]; yet not WN-like), which is modified by the algorithm so that its synsets are not the fuzzy version of the previous synsets. The aforementioned studies have not received much attention from the text mining community, whose research efforts utilize platforms defined on already existing WLDs. The community is reluctant to change its foundational platforms and migrate to, although useful, different and new ones. This is the reason, in our opinion, why fuzzy synsets are kept almost isolated in the field of Text Mining. To the best of our knowledge no research[7] has solved this shortcoming; León-Araúz et al. (yet from their fuzzy-ontology viewpoint), mention it alongside their study: "extending WordNet and EuroWordNet to include imprecise knowledge requires a considerable effort to define synset membership, similarity and equivalence degrees;" however, they did not propose any approach.

In the position paper version of this study [23], in 2016, we have proposed an idea for overcoming this drawback which is going to be described in details, extended, and implemented in this paper.

In this paper, we present an algorithm able to assign membership functions for predefined synsets of any language, given a large corpus of documents of that language and a Word Sense Disambiguation (WSD)[8] as input. Then, we apply the algorithm to the English language, using the Open American National Corpus (OANC) and the well-known graph-based WSD system, UKB, and construct the fuzzy version of WordNet, (FWN) accessible online.

Section 2 introduces our algorithm, able to produce the Fuzzified WLD of any language, theoretically proves the validity of its output, and describes the online version of the proposed FWN and section 3 ends the paper with conclusions and future directions.

## II. PRODUCING FUZZY SYNSETS FOR PREDEFINED SYNSETS

In this section, we propose an algorithm for constructing fuzzy synsets in any language. As its input, the algorithm requires: (1) A large corpus ($C$) of documents of that language and (2) a WSD algorithm $W$ (each WSD algorithm is paired with a WLD and each WLD contains a set $S^{(W)}$ of synsets of that language; $u_{i,k}^{(W)}$ stands for the word-sense $k$ from the synset $i$ of the WLD engaged with $W$, and $S_k^{(W)}$ stands for the $k^{th}$ synset of $W$.).

This algorithm is comprised of the following 4 steps:

**Frequency:** For each word-sense $u_{i,k}^{(W)}$ of each synset $S_k^{(W)}$ calculate $f^{(C,W)}(u_{i,k}^{(W)})$, that is the frequency of $u_{i,k}^{(W)}$ in $C$.

**Probability:** For each word-sense $u_{i,k}^{(W)}$ of each synset $S_k^{(W)}$ calculate

$$pr^{(C,W)}(u_{i,k}^{(W)}) = f^{(C,W)}(u_{i,k}^{(W)}) / \Sigma_{u_{m,k} \in S_k} f^{(C,W)}(u_{m,k}^{(W)}).$$

**Possibility:** For each word-sense $u_{i,k}^{(W)}$ of each synset $S_k^{(W)}$ calculate

$$\pi_{1983}^{(C,W)}(u_{i,k}^{(W)}) = \Sigma_{u_{m,k}^{(W)} \in S^{(W)}} \min\left(pr^{(C,W)}(u_{i,k}^{(W)}), pr^{(C,W)}(u_{m,k}^{(W)})\right)$$

$$\pi_{1993}^{(C,W)}(u_{i,k}^{(W)}) = \Sigma_{u_{m,k}^{(W)} | pr^{(C,W)}(u_{m,k}^{(W)}) \leq pr^{(C,W)}(u_{i,k}^{(W)})} pr^{(C,W)}(u_{m,k}^{(W)})$$

**Membership:** For each word-sense $u_{i,k}^{(W)}$ of each synset $S_k^{(W)}$ calculate the membership degree of $u_{i,k}^{(W)}$ in the fuzzy set $S_k^{(W)}$

$$\mu_{S_k,1983}^{(C,W)}(u_{i,k}^{(W)}) = \pi_{1983}^{(C,W)}(u_{i,k}^{(W)})$$
$$\mu_{S_k,1993}^{(C,W)}(u_{i,k}^{(W)}) = \pi_{1993}^{(C,W)}(u_{i,k}^{(W)}).$$

### A. Proof of the algorithm

Here, we propose a theoretical proof, for the algorithm validity.

$pr^{(C,W)}() =$ **probability.**

**Definition 1.** Given a WSD algorithm W and a corpus of ordered documents $C$, the sequence $L_{k,C,W} = \left(l_{k,C,W}^{(a)}\right)_{a=1}^{n}$, is defined so that $l_{k,C,W}^{(a)}$ represents the $a^{th}$ occurrence of any of the word-senses (recognized by $W$) of the synset $S_k$ in $C$.

**Definition 2.** For a WSD $W$, $U_{i,k,W}: S_k^{(W)} \to \{0,1\}$ is defined as a Bernoulli random variable that for a given $u \in S_k^{(W)}$, it outputs 1 if $u = u_{i,k}^{(W)}$ and outputs 0, otherwise.

**Definition 3.** The Bernoulli process $C_{i,k,W}$ is defined as the sequence of random variables $\left\{U_{i,k,C,W}^{(a)}\right\}_{a=1}^{|L_{k,C,W}|}$, which its $a^{th}$

---

[4] For more information about synonymity please refer to [40]

[5] A full-scale Swedish lexical-semantic resource with non-classical, associative relations among word and multiword senses, identified by persistent formal identifiers.

[6] the original reference was older version of [22]

[7] There is a similar concept not to be confused with this discussion that is "graded word sense assignment" [41] that addresses fine-grained graded versions of word-senses of lemmas whereas we are addressing fuzzy synsets (fine-grained graded versions of word-senses of synsets).

[8] In cognitive and computational linguistics, Word Sense Disambiguation (WSD) is an open problem belonging to ontology and natural language processing. Considering a word in a sentence, WSD identifies which of its senses is used in that sentence (for multi-sense words) [42].

element represents $U_{i,k,W}(l_{k,C,W}^{(a)})$, for an arbitrary corpus $C$ and WSD $W$.

**Lemma 1.** Consider an arbitrary Bernoulli process $C_{i,k,W}$, assuming that $\{U_{i,k,C,W}^{(a)}\}_{a=1}^{|L_{k,C,W}|}$ are independent and identically distributed (i.i.d) Bernoulli random variables with success probability of $pr_{i,k}$. Then, for the random variable $\overline{U_{i,k,C,W}} = \frac{1}{|L_{k,C,W}|}\sum_{i=1}^{|L_{k,C,W}|} U_{i,k,C,W}^{(a)}$, we have $Pr\left(\lim_{|L_{k,C,W}|\to\infty}\overline{U_{i,k,C,W}} = P(U_{i,k,W} = u_{i,k}|S_k)\right) = 1$.

**Proof.** A direct result of the Khintchine's Strong Law of Large Numbers [24] results in $Pr\left(\lim_{|L_{k,C,W}|\to\infty}\overline{U_{i,k,C,W}} = pr_{i,k}\right)$. Moreover, we know that $\forall a \in \{1,2,\dots,|L_{k,C,W}|\}: pr_{i,k} = pr_{i,k}^{(a)} = P(U_{i,k,C,W}^{(a)} = u_{i,k}|S_k)$. However, we know that the i.i.d $U_{i,k,C,W}^{(a)}$ Bernoulli random variables are the i.i.d elements of the Bernoulli process $C_{i,k,W}$. This implies that $\forall a \in \{1,2,\dots,|L_{k,C,W}|\}: U_{i,k,C,W}^{(a)} = U_{i,k,W}(l_{k,C,W}^{(a)})$. In other words, $U_{i,k,C,W}^{(a)}$ are tantamount to i.i.d trials of the random variable $U_{i,k,W}$, all of which having the distribution $U_{i,k,W}$. Thus, we can write $pr_{i,k} = pr_{i,k}^{(a)} = P(U_{i,k,C,W}^{(a)} = u_{i,k}|S_k) = P(U_{i,k,W} = u_{i,k}|S_k)$. ∎

**Definition 4.** Given a WSD algorithm W and a corpus of ordered documents $C$, the sequence $L_{C,W} = (l_{C,W}^{(a)})_{a=1}^{n}$, is defined so that $l_{C,W}^{(a)}$ represents the $a^{th}$ occurrence of any of the word-senses (recognized by $W$) in $C$.

**Definition 5.** For a WSD $W$, $U_{k,W}: WLD(W) \to \{0,1\}$ is defined as a Bernoulli random variable that for a given $u \in WLD(W)$, it outputs 1 if $u \in S_k$ and outputs 0, otherwise, where $WLD(W)$ stands for the WLD, engaged with the WSD $W$.

**Definition 6.** The Bernoulli process $C_{k,W}$ is defined as the sequence of random variables $\{U_{k,C,W}^{(a)}\}_{a=1}^{|L_{C,W}|}$, which its $a^{th}$ element represents $U_{k,W}(l_{C,W}^{(a)})$, for an arbitrary corpus $C$ and WSD $W$.

**Lemma 2.** In an arbitrary Bernoulli process $C_{k,W}$, assuming that $\{U_{k,C,W}^{(a)}\}_{a=1}^{|L_{C,W}|}$ are i.i.d Bernoulli random variables with success probability of $pr_k = P(U_{k,W} \in S_k)$, then, for the random variable $\overline{U_{k,C,W}} = \frac{1}{|L_{C,W}|}\sum_{i=1}^{|L_{C,W}|} U_{k,C,W}^{(a)}$, we have $Pr\left(\lim_{|L_{C,W}|\to\infty}\overline{U_{k,C,W}} = pr_k\right) = 1$.

**Proof.** The same as the proof of Lemma 1. ∎

**Lemma 3.** Consider an arbitrary infinitely-large corpus $C$, a precise WSD $W$, and a probable $S_k$. If the usage of each word-sense / synset, in $C$, is independent of the usage of other word-senses / synsets, we almost surely, have $|L_{k,C,W}| \to +\infty$.

**Proof.** Because $C$ is infinitely large ($|L_{C,W}| \to +\infty$), Lemma 2 implies that $Pr\left(\sum_{i=1}^{|L_{C,W}|} U_{k,C,W}^{(a)} = pr_k \cdot |L_{C,W}|\right) = 1$. But, we know that $S_k$ is probable (i.e. $pr_k > 0$), and therefore, $\sigma_k = pr_k \cdot |L_{C,W}| \to +\infty$. Moreover, we know that $|L_{k,C,W}| = \sum_{i=1}^{|L_{C,W}|} U_{k,C,W}^{(a)}$. Thus, we have $Pr(|L_{k,C,W}| = \lim_{\sigma_k \to +\infty} \sigma_k) = 1$. Thus, almost surely, $|L_{k,C,W}| \to +\infty$. ∎

**Theorem 1.** Consider an arbitrary infinitely-large corpus $C$, a precise WSD $W$, and a probable $S_k$. If the usage of each word-sense / synset, in $C$, is independent of the usage of other word-senses / synsets, we almost surely, have $pr_{i,k} = pr^{(C,W)}(u_{i,k}^{(W)}|S_k) = f^{(C,W)}(u_{i,k}^{(W)})/\sum_{u_{m,k}\in S_k} f^{(C,W)}(u_{m,k}^{(W)})$.

**Proof.** Lemma 3 implies that $|L_{k,C,W}| \to +\infty$. Now, Lemma 1 implies that for any $u_{i,k} \in S_k$, we have $Pr\left(\frac{1}{|L_{k,C,W}|}\sum_{i=1}^{|L_{k,C,W}|} U_{i,k,C,W}^{(a)} = P(U_{i,k,W} = u_{i,k}|S_k)\right) = 1$. However, we know that $\sum_{i=1}^{|L_{k,C,W}|} U_{i,k,C,W}^{(a)} = f^{(C,W)}(u_{i,k}^{(W)})$ and also know that $|L_{k,C,W}| = \sum_{u_{m,k}\in S_k} f^{(C,W)}(u_{m,k}^{(W)})$. Therefore, we have $Pr\left(\frac{1}{\sum_{u_{m,k}\in S_k} f^{(C,W)}(u_{m,k}^{(W)})}\sum_{i=1}^{|L_{k,C,W}|} f^{(C,W)}(u_{i,k}^{(W)}) = P(U_{i,k,W} = u_{i,k}|S_k)\right) = 1$, and equally, $Pr\left(pr^{(C,W)}(u_{i,k}^{(W)}) = P(U_{i,k,W} = u_{i,k}|S_k)\right) = 1$. ∎

## $\pi(u_{i,k})$ = possibility.

**Definition 7** [25]. The degree of necessity of event $A \subseteq X$ is the extra amount of probability of elementary events in $A$ over the amount of probability assigned to the most frequent elementary event outside $A$. In other words, $N(A)$ is defined as the necessity measure of $A$, so that, $N(A) = \sum_{x_i \in A} \max(pr_i - \max_{x_k \notin A} pr_k)$. It is also called the Shafer's consonant belief function [26].

**Preposition 1.** $N(A)$ satisfies the following 3 axioms of necessity function: $N(\emptyset) = 0$, $N(X) = 1$, and $\forall A, B \subseteq X, N(A \cap B) = \min(N(A), N(B))$.

**Proof.** proven in [25]. ∎

**Definition 8** [25]. "Viewing $N(A)$ as the grade of impossibility of the opposite event $\bar{A}$ we can define the grade of possibility of $A$ by $\forall A \subseteq X, \Pi(A) = 1 - N(\bar{A})$."

**Preposition 2.** The set function $\Pi$ is a possibility measure in the sense of Zadeh [27].

**Proof.** proven in [25]. ∎

**Lemma 4.** Consider $\pi(x), pr(x)$ as possibility and probability mass functions, engaged with the Possibility and Probability distributions $\Pi$ and $P$. Adopting the Shafer's consonant belief function as the necessity measure, we will have $\pi(x_i) = \sum_{j=1}^{n} \min(pr(x_i), pr(x_j)), \forall x_i \in X$.

**Proof.** proven in [25]. ∎

**Theorem 2.** Consider an arbitrary infinitely-large corpus $C$, a precise WSD $W$, and a probable $S_k$. If the usage of each word-sense / synset, in $C$, is independent of the usage of other word-senses / synsets, and if the Shafer's consonant belief function is adopted as the necessity measure, then, for any $u_{i,k} \in S_k$, we almost surely, will have $\pi_{i,k} = \pi_{1983}^{(C,W)}(u_{i,k}^{(W)}) = \sum_{u_{m,k}^{(W)} \in S_k^{(W)}} \min(pr^{(C,W)}(u_{i,k}^{(W)}), pr^{(C,W)}(u_{m,k}^{(W)}))$.

**Proof.** Theorem 1 implies that, almost surely,

$pr^{(C,W)}(u_{i,k}^{(W)}) = P(U_{i,k,W} = u_{i,k}|S_k)$. Using this fact, besides Lemma 4, we almost surely will have

$\pi(u_{i,k}|S_k) = \sum_{j=1}^{n} \min(pr^{(C,W)}(u_{i,k}^{(W)}), pr^{(C,W)}(u_{j,k}^{(W)})) =$
$\sum_{u_{m,k}^{(W)} \in S_k^{(W)}} \min(pr^{(C,W)}(u_{i,k}^{(W)}), pr^{(C,W)}(u_{m,k}^{(W)}))$, or equally, $\pi_{1983}^{(C,W)}(u_{i,k}^{(W)})$. Therefore, we almost surely have $\pi(u_{i,k}|S_k) = \pi_{1983}^{(C,W)}(u_{i,k}^{(W)})$. ∎

**Definition 9** [28]**.** Consider the probability distribution $P$ and possibility distribution $\Pi$ defined on $X$. Then, $P$ and $\Pi$ have DP-consistency[9] if $\forall A \subseteq X, P(A) \leq \Pi(A)$.

**Proposition 3.** DP-consistency is a standard consistency measure in the sense of Delgado-Moral.

**Proof.** Proven in [29].

**Definition 10.** Consider the probability distribution $P$ and possibility distribution $\Pi$, defined on $X$. Then, $P$ and $\Pi$ have the preference-preservation relation if $\forall x, x' \in X: \pi(x) > \pi(x') \Leftrightarrow pr(x) > pr(x')$, where $\pi(x)$ and $pr(x)$ are the possibility and probability mass functions, engaged with $\Pi$ and $P$, both defined on $X \to [0,1]$.

**Preposition 4.** The condition $\forall x, x' \in X: \pi(x) > \pi(x') \Leftrightarrow pr(x) > pr(x')$ is equal with $\pi(x) < \pi(x') \Leftrightarrow pr(x) < pr(x')$ or $\pi(x) \leq \pi(x') \Leftrightarrow pr(x) \leq pr(x')$ or $\pi(x) \geq \pi(x') \Leftrightarrow pr(x) \geq pr(x')$.

**Proof.** Considering that $x$ and $x'$ do not have any discriminative specificity, the condition can be read as $\pi(x') < \pi(x) \Leftrightarrow pr(x') < pr(x)$. Moreover, contraposition of the mentioned equal conditions, yields in conditions with $\leq$ and $\geq$.

**Definition 11** [30]**.** Given $X$ as a finite set of elements and $P, \Pi$ as probability and possibility distributions on $X$, and $p, \pi$ the corresponding mass functions, the transformed possibility $\pi$ is maximally specific when $\sum_{x \in X} \pi(x)$ has the minimum value, respecting preference-preservation and DP-consistency of $P, \Pi$.

**Lemma 5.** Given a probability distribution $P$ and probability mass function $pr(x)$ in the finite Universe of discourse $X$, the possibility distribution $\Pi$, in the same time, satisfies the 3 restrictions: DP-consistency, preference preservation, and maximally specificity, if and only if $\forall x \in X, \pi(x) = \sum_{\{x': pr(x') \leq pr(x)\}} pr(x')$.

**Proof.** Without losing the generality, suppose that $X = \{x_1, x_2, \dots, x_n\}$ while (upon Preposition 4) we have $pr(x_1) \leq pr(x_2) \leq \dots \leq pr(x_n)$. Utilizing Preposition 4, preference preservation implies that $\pi(x_1) \leq \pi(x_2) \leq \dots \leq \pi(x_n)$. Consider $A_i = \{x_1, x_2, \dots x_i\}$. DP-consistency implies that $\forall A_i, \Pi(A_i) \geq P(A_i)$. Thus, $\forall A_i, \max\{\pi(x_1), \pi(x_2), \dots \pi(x_i)\} \geq \sum_{k=1}^{i} pr(x_k)$. Therefore, we have $\forall A_i, \pi(x_i) \geq \sum_{k=1}^{i} pr(x_k)$. Now, because $\pi(x_i) = \sum_{k=1}^{i} pr(x_k)$, from the one hand satisfies the preference preservation and DP-consistency restrictions, and from the other hand, includes the minimum allowed values of the $\pi(x_i) \geq \sum_{k=1}^{i} pr(x_k)$ constraint, $\pi(x_i) = \sum_{k=1}^{i} pr(x_k)$ would be the unique minimal case satisfying the 3 mentioned constraints. Please note that the expressions $\pi(x_i) = \sum_{k=1}^{i} pr(x_k)$ and $pr(x_1) \leq pr(x_2) \leq \dots \leq pr(x_n)$ equals with $\pi(x_i) = \sum_{\{x_k: pr(x_k) \leq pr(x_i)\}} pr(x_k)$. ∎

Please note that the formula $\pi(x_i) = \sum_{\{x_k: pr(x_k) \leq pr(x_i)\}} pr(x_k)$ although introduced in 1982 [31], it is usually known and referenced by [30], a better known research work from 1993 where the same authors propose both its discrete and continuous versions.

**Theorem 3.** Consider an arbitrary infinitely-large corpus $C$, a precise WSD $W$, and a probable $S_k$. If the usage of each word-sense / synset, in $C$, is independent of the usage of other word-senses / synsets, and if the 3 constraints of DP-consistency, preference-preservation, and maximally specificity have to be satisfied, then, for any $u_{i,k} \in S_k$, we almost surely, will have $\pi_{i,k} = \pi_{1993}^{(C,W)}(u_{i,k}^{(W)}) =$
$\sum_{u_{m,k}^{(W)} | pr^{(C,W)}(u_{m,k}^{(W)}) \leq pr^{(C,W)}(u_{i,k}^{(W)})} pr^{(C,W)}(u_{m,k}^{(W)})$.

**Proof.** Theorem 1 implies that, almost surely, $pr^{(C,W)}(u_{i,k}^{(W)}) = P(U_{i,k,W} = u_{i,k}|S_k)$. Using this fact, besides the Lemma 5, we almost surely will have $\pi(u_{i,k}|S_k) = \sum_{u_{m,k}^{(W)} | pr^{(C,W)}(u_{m,k}^{(W)}) \leq pr^{(C,W)}(u_{i,k}^{(W)})} pr^{(C,W)}(u_{m,k}^{(W)})$, or equally, $\pi_{1993}^{(C,W)}(u_{i,k}^{(W)})$. Therefore, we almost surely have $\pi(u_{i,k}|S_k) = \pi_{1993}^{(C,W)}(u_{i,k}^{(W)})$. ∎

Please note that Dubois and Prade, in 1993 [30], illustrate that the possibility mass function v.93 provides a maximally informative transformation from probability to possibility distribution. Both transformations have advantages and drawbacks; the possibility mass function v.83 produces more homogeneous values, denser around 1 and always greater than or equal to v.93 values[10]. However, being the v.93 the maximally informative transformation, we expect the Fuzzified WLDs v.93 to be more efficient (than v.83) in Text-Mining applications.

$$\mu_{S_k}(u_{i,k}) = \pi(u_{i,k}).$$

**Definition 12** [27]**.** Let $F$ be a fuzzy subset of a universe of discourse $U$, which is characterized by its membership function $\mu_F$, with the grade of membership, $\mu_F(u)$, interpreted as the compatibility of $u$ with the concept labeled $F$. Also, Let $X$ be a variable taking values in $U$. Then, $F$ is postulated to act as a fuzzy restriction, $R(X)$, associated with $X$ and the proposition "$X$ is $F$," translates into $R(X) = F$.

**Definition 13** [27]**.** An arbitrary fuzzy restriction $R(X)$ associates a possibility distribution, $\Pi_X$, with $X$ which is postulated to be equal to $R(X)$ (i.e., $\Pi_X = R(X)$).

**Definition 14** [27]**.** Consider a fuzzy set $F$, a variable $X$ taking values in the universe of discourse $U$ and the $R(X)$ associated with $F$ and $X$. The possibility distribution function associated with $X$ is denoted by $\pi_X$ and is defined to be numerically equal to the membership function of F (i.e. $\pi_X \triangleq \mu_F$).

**Lemma 6.** Consider a fuzzy set $F$, a variable $X$ taking

---

[9] DP stands for Dubois-Prade. There are two other consistency measures, proposed by Zadeh [27] and Sugeno [43]. The interested reader is referred to Delgado and Moral [29] which analyzes these three, in detail.

[10] The least informative version is a version that assigns 1 to possibility of the entire classes.

5values in the universe of discourse $U$, and the $\Pi(X)$, associated with $F$ and $X$. Then, $\pi_X(u)$ the possibility that $X = u$, given that "X is F," is postulated to be equal to $\mu_F(u)$.

**Proof.** Upon Definition 12, we know that "$X$ is $F$," translates into $R(X) = F$ and upon Definition 13, we know that $\Pi_X = R(X)$. Thus, "$X$ is $F$," is an intrinsic assumption in $\Pi_X$. Moreover, upon the Definition 14, it is postulated that $\pi_X \triangleq \mu_F$. Thus, $\pi_X(u) = \mu_F(u)$, given that "X is F." In other words $\pi_X(u)$ equals the possibility that $X = u$, given that "X is F." ∎

**Lemma 7.** Consider a fuzzy synonym-set (synset) $S_k$, a variable $U_{i,k,W}$ taking values in the universe of discourse $WLD(W)$, and the $\Pi(U_{i,k,W})$, associated with $S_k$ and $U_{i,k,W}$. Then, $\pi_{U_{i,k,W}}(u_{i,k})$ the possibility that $U_{i,k,W} = u_{i,k}$, given that "$U_{k,C,W}$ is in $S_k$," is postulated to be equal to $\mu_{S_k}(u_{i,k})$ (i.e. $\pi(U_{i,k,W} = u_{i,k}|S_k) \triangleq \mu_{S_k}$).

**Proof.** A direct result of them Lemma 6. ∎

**Theorem 4.** Consider an arbitrary infinitely-large corpus $C$, a precise WSD $W$, and a probable $S_k$. If in $C$, the usage of each word-sense / synset is independent of the usage of other word-senses / synsets.

(a) If the Shafer's consonant belief function is adopted as the necessity measure, then, for any $u_{i,k} \in S_k$, we almost surely, will have $\mu_{S_k} = \pi_{1983}^{(C,W)}(u_{i,k}^{(W)})$.

(b) If the 3 constraints of DP-consistency, preference-preservation, and maximally specificity have to be satisfied, then, for any $u_{i,k} \in S_k$, we almost surely, will have $\mu_{S_k} = \pi_{1993}^{(C,W)}(u_{i,k}^{(W)})$.

**Proof.** (a) By Theorem 2, given the assumptions of part (a), we would have $\pi(u_{i,k}|S_k) = \pi_{1983}^{(C,W)}(u_{i,k}^{(W)})$. Also the Lemma 7 implies that $\mu_{S_k} \triangleq \pi(U_{i,k,W} = u_{i,k}|S_k)$. Thus, we have $\mu_{S_k} = \pi_{1983}^{(C,W)}(u_{i,k}^{(W)})$.

(b) By Theorem 3, given the assumptions of part (b), we have $\pi(u_{i,k}|S_k) = \pi_{1993}^{(C,W)}(u_{i,k}^{(W)})$. Also the Lemma 7 implies that $\mu_{S_k} \triangleq \pi(U_{i,k,W} = u_{i,k}|S_k)$. Thus, we have $\mu_{S_k} = \pi_{1993}^{(C,W)}(u_{i,k}^{(W)})$. ∎

*B. Pseudocode of the algorithm*

In the following, you see the pseudocode of the algorithm.

*//WSF is Words-Sense Frequency matrix. The 1st dimension is for synsets and the 2nd dimension is for its word-senses. Each cell represents the frequency of a synset's member (word-sense) in the whole corpus.*
*//PMV, Possibility1983, and Possibility1993 stand for Probability Mass Value, Possibility (v.83) mass value, and Possibility (v.93) mass value, respectively. Dimensions are the same as what in WSF.*

```
For i = 1 to total number of synsets
  synSize = numberOfWord-senses(synset[i]);
  totalFrequencyOfSynset = synSize;
  For j = 1 to synSize
    totalFrequencyOfSynset += WSF[i][j];
  For j = 1 to synSize
    PMV[i][j] = (WSF[i][j]+1) /
            totalFrequencyOfSynset;
  For j = 1 to synSize
    possibility1983OfJ = 0;
    possibility1993OfJ = 0;
    pIJ = PMV[i][j];
    For m = 1 to synSize
      pIM = PMV[i][m];
      possibility1983OfJ += min(pIJ,pIM);
      possibility1993OfJ +=
              piecewise(pIM <= pIJ , pIM , 0);
    Possibility1983[i][j] = possibility1983OfJ;
    Possibility1993[i][j] = possibility1993OfJ;
```

Please note that the above pseudocode utilizes the auxiliary technique of smoothing for bypassing the realistic limitations, occurring when the frequency of some word-senses in the corpus in zero. This is the reason why totalFrequencyOfSynset is initialized by synSize, as it is assumed that each word-sense of a synset is visited once before analysing the corpus.

Although we proposed a proof for the validity of the results of the abovementioned algorithm, considering the real-world experiments-limitations, the above pseudocode produces the accurate membership values of the predefined synsets of the lexical database associated with the utilized WSD algorithm if and only if two conditions are satisfied.

**Condition 1:** Corpus is large enough to provide accurate probability values, as a basis for membership functions. This is because the corpus has to be large enough to satisfy the law of large numbers (utilized in the first step of the proof).

**Condition 2:** WSD algorithm works precisely so that the recognized word-senses will be trustable. This is because the $f(u_{i,k})$ function is fed by the output of WSD algorithm and if it does not work properly, the results in all the next steps will be corrupted.

*C. Applying the algorithm for fuzzification of the standard WordNet*

To apply our algorithm to the English language, as the algorithm input we use the English corpus "Open American National Corpus" (OANC [32], comprising almost 16.6 million words [32][33]) and the well-known graph-based Word Sense Disambiguation algorithm "UKB." We publish the entire list of English fuzzy synsets for both versions (v.83 and v.93) online. It can be found at http://bayanbox.ir/info/1272736121331182587/fuzzy-synsets.

About competence of UKB for our algorithm, satisfying the abovementioned second condition (WSD precision), it is worthy to note that the UKB has been evaluated in several outstanding research tasks including usage of WN for WSD [34][35], WSD on medical domain [36], improvements of Information Retrieval using WN [37][38], Word Embedding[11] on WN [39], etc. It is also worthy to remind that the proposed algorithm (for producing fuzzy synsets) is language-free and the interested researcher can apply it to his favorite language.

III. CONCLUSIONS AND FUTURE WORK

In this study, we propose an algorithm for the automatic generation of fuzzy membership functions for definite synsets of the existing WordNet-like Lexical Databases (WLDs). The proposed WLD-fuzzifier algorithm is mainly based on the definition of possibility and its relationship with membership functions, and also, the validity of its results is proven, mathematically, by the Probability and Possibility Theorem methods. Moreover, we apply the proposed algorithm to the English language to generate the fuzzified version of WordNet (FWN) and publish it online. As a future trend of this study, the WLD-fuzzifier algorithm proposed in this paper is recommended to fuzzify every other WLD in any language to increase the Text-Mining efficiency in those languages.

---

[11] Produced with random walk

REFERENCES

[1] G. a. Miller, R. Beckwith, C. Fellbaum, D. Gross, and K. J. Miller, "Introduction to wordnet: An on-line lexical database," *Int. J. Lexicogr.*, vol. 3, no. 4, pp. 235–244, 1990.

[2] G. a. Miller, "WordNet: a lexical database for English," *Commun. ACM*, vol. 38, no. 11, pp. 39–41, 1995.

[3] C. Fellbaum, "WordNet: An Electronic Lexical Database," *Br. J. Hosp. Med. London Engl. 2005*, vol. 71, no. 3, p. 423, 1998.

[4] F. Bond and K. Paik, "A Survey of WordNets and their Licenses," in *Proceedings of the 6th Global WordNet Conference (GWC 2012)*, 2012, pp. 64–71.

[5] P. Vossen, "Introduction to EuroWordNet," in *Computers and the Humanities*, 1998, vol. 32, pp. 73–89.

[6] P. Vossen, "EuroWordNet: A multilingual database of autonomous and language-specific WordNets connected via an inter-lingual-index," *Int. J. Lexicogr.*, vol. 17, no. 2, pp. 161–173, 2004.

[7] T. Wei, Q. Zhou, H. Chang, Y. Lu, and X. Bao, "A Semantic Approach for Text Clustering using WordNet and Lexical Chains," *Expert Syst. Appl.*, 2015.

[8] D. Reforgiato Recupero, V. Presutti, S. Consoli, A. Gangemi, and A. G. Nuzzolese, "Sentilo: Frame-Based Sentiment Analysis," *Cognit. Comput.*, vol. 7, no. 2, pp. 211–225, 2015.

[9] J. Yan, C. Wang, W. Cheng, M. Gao, and A. Zhou, "A retrospective of knowledge graphs," *Front. Comput. Sci.*, vol. 12, no. 1, pp. 55–74, 2018.

[10] M. Ivasic-Kos, M. Pobar, and S. Ribaric, "Two-tier image annotation model based on a multi-label classifier and fuzzy-knowledge representation scheme," *Pattern Recognit.*, vol. 52, pp. 287–305, 2016.

[11] D. Madalli, A. Sulochana, and A. K. Singh, "COMAT: core ontology of matter," *Program*, vol. 50, no. 1, pp. 103–117, 2016.

[12] K. D. Bimson, R. D. Hull, and D. Nieten, "The Lexical Bridge: A Methodology for Bridging the Semantic Gaps between a Natural Language and an Ontology," in *Semantic Web*, 2016, pp. 137–151.

[13] E. Velldal, "A fuzzy clustering approach to word sense discrimination," in *Proceedings of the 7th International conference on Terminology and Knowledge Engineering*, 2005.

[14] L. Borin and M. Forsberg, "From the people's synonym dictionary to fuzzy synsets-first steps," in *Proc. LREC 2010 workshop Semantic relations. Theory and Applications.*, 2010.

[15] J. Hurford, "Why Synonymy is Rare: Fitness is in the Speaker," *Ecal03*. pp. 442–451, 2003.

[16] L. Borin and M. Forsberg, "Beyond the synset: Swesaurus–a fuzzy Swedish wordnet," in *Proceedings of the symposium: Re-thinking synonymy: semantic sameness and similarity in languages and their description*, 2010.

[17] V. Kann and M. Rosell, "Free construction of a free Swedish dictionary of synonyms," in *Proc. 15th Nordic Conf. on Comp. Ling.–NODALIDA (5)*, 2005, pp. 1–6.

[18] L. Borin, "Mannen är faderns mormor: Svenskt associationslexikon reinkarnerat," *LexicoNordica*, vol. 12, pp. 39–54, 2005.

[19] L. Borin and M. Forsberg, "All in the family: A comparison of SALDO and WordNet," in *Proceedings of the Nodalida 2009 Workshop on WordNets and other Lexical Semantic Resources - between Lexical Semantics, Lexicography, Terminology and Formal Ontologies. NEALT Proceedings Series*, 2009.

[20] H. Gonçalo Oliveira and P. Gomes, "Automatic Discovery of Fuzzy Synsets from Dictionary Definitions," in *22nd International Joint Conference on Artificial Intelligence*, 2011, pp. 1801–1806.

[21] A. Kilgarriff, "'I don't believe in word senses,'" *Comput. Hum.*, vol. 31, no. 2, p. 25, 1997.

[22] G. Hirst, "Ontology and the Lexicon," in *Ontology and the Lexicon*, S. Staab and R. Studer, Eds. Springer Berlin Heidelberg, 2009, pp. 269–292.

[23] S.-A. Hossayni, M. R. Akbarzadeh-T, D. Reforgiato Recupero, A. Gangemi, and J. L. de la Rosa i Esteva, "Fuzzy Synsets, and Lexicon-Based Sentiment Analysis," in *EMSA-RMed*, vol. 1613, D. M. Deng Y., Denecke K., Declerck T., Recupero D.R., Ed. CEUR Workshop Proceedings (Scopus code 21100218356), 2016.

[24] P. K. Sen and J. M. Singer, *Large Sample Methods in Statistics: An Introduction with Applications*. CRC Press, 1993.

[25] D. Dubois and H. Prade, "Unfair coins and necessity measures: Towards a possibilistic interpretation of histograms," *Fuzzy Sets Syst.*, vol. 10, no. 1–3, pp. 15–20, 1983.

[26] G. Shafer, *A Mathematical Theory of Evidence*. Princeton university press, 1976.

[27] L. a. Zadeh, "Fuzzy Sets as a basis for a theory of possibility," *Fuzzy Sets Syst.*, vol. 1, no. 1, pp. 3–28, 1978.

[28] D. J. Dubois and H. Prade, *Fuzzy Sets and Systems: Theory and Applications*. 1980.

[29] M. Delgado and S. Moral, "On the concept of possibility-probability consistency," *Fuzzy Sets Syst.*, vol. 21, no. 3, pp. 311–318, 1987.

[30] D. Dubois, H. Prade, and S. Sandri, "On possibility/probability transformations," *Proc. Fourth IFSA Conf.*, pp. 103–112, 1993.

[31] D. Dubois and H. Prade, "On several representations of an uncertain body of evidence," in *Fuzzy information and decision processes*, 1982, pp. 167–181.

[32] C. J. Fillmore, N. Ide, D. Jurafsky, and C. Macleod, "An American National Corpus: A Proposal," in *the First International Language Resources and Evaluation Conference*, 1998, pp. 965–970.

[33] G. de Melo, C. F. Baker, N. Ide, R. J. Passonneau, and C. Fellbaum, "Empirical Comparisons of MASC Word Sense Annotations," in *Lrec 2012 - Eighth International Conference on Language Resources and Evaluation*, 2012, pp. 3036–3043.

[34] E. Agirre and A. Soroa, "Personalizing PageRank for Word Sense Disambiguation," in *Proceedings of the 12th Conference of the European Chapter of the ACL*, 2009, no. April, pp. 33–41.

[35] S. A. Agirre E., Lopez de Lacalle O., "Random Walks for Knowledge-Based Word Sense Disambiguation," *Comput. Linguist. 401*, vol. 40, no. 1, 2014.

[36] D. Martinez, A. Otegi, A. Soroa, and E. Agirre, "Improving search over Electronic Health Records using UMLS-based query expansion through random walks," *J. Biomed. Inform.*, vol. 51, pp. 100–106, 2014.

[37] A. Otegi, A. Xavier, and A. Eneko, "Query Expansion for IR using Knowledge-Based Relatedness," in *Proceedings of 5th International Joint Conference on Natural Language Processing*, 2011, pp. 1467–1471.

[38] A. Otegi, X. Arregi, O. Ansa, and E. Agirre, "Using knowledge-based relatedness for information retrieval," *Knowl. Inf. Syst.*, vol. 44, no. 3, pp. 689–718, 2014.

[39] J. and E. A. and A. S. Goikoetxea, "Random Walks and Neural Network Language Models on Knowledge Bases," in *Proceedings of the North American Chapter of the Association for Computational Linguistics: Human Language Technologies*, 2015, pp. 1434–1439.

[40] C. E. Osgood, "the Nature and Measurement of Meaning," *Psychol. Bull.*, vol. 49, no. 3, p. 227, 1952.

[41] K. Erk, D. McCarthy, and N. Gaylord, "Measuring word meaning in context," *Comput. Linguist.*, vol. 39, no. 3, pp. 511–554, 2013.

[42] W. Weaver, "Translation," *Mach. Transl. Lang.*, vol. 14, pp. 15–23, 1955.

[43] M. Sugeno, "Theory of fuzzy integrals and its application," 1972.